\documentclass{article}
\usepackage{spconf,amsmath,graphicx,hyperref}

\usepackage{booktabs}
\usepackage{xcolor}
\definecolor{greenhighlight}{RGB}{230, 245, 230}
\definecolor{grayhighlight}{RGB}{235, 235, 235}
\definecolor{lightgray2}{RGB}{247, 247, 247}
\definecolor{lightgray3}{RGB}{241, 241, 241}
\definecolor{darkgreen}{RGB}{34, 139, 34}
\definecolor{darkred}{RGB}{139,0,0}

\usepackage{cite}
\usepackage{amsmath}
\usepackage{amssymb}
\usepackage{amsfonts}
\usepackage{bbm}
\usepackage{graphicx}
\usepackage{subcaption}
\usepackage{balance}
\usepackage{amssymb}
\usepackage{pifont}

\usepackage{array}
\usepackage{color, xcolor}
\usepackage{colortbl}
\usepackage{booktabs}
\usepackage{multicol}
\usepackage{multirow}
\usepackage{adjustbox}

\usepackage{enumitem}

\newcommand{\modelname}{BGC-Nav}

\title{\modelname{}: Cross-Altitude Multi-UAV Collaborative Navigation via Belief-Guided Coupling}
\name{Jiuwu Hao$^{1,2,*}$,
Yueyang Wu$^{1,2,*}$,
Liguo Sun$^{1}$,
Haolin Song$^{1,2}$,
Zao Zhang$^{1, \dagger}$,
Pin Lv$^{1,\dagger}$
\thanks{$^{*}$~Equal contribution. \quad $^{\dagger}$~Corresponding authors.}}
\address{$^{1}$Institute of Automation, Chinese Academy of Sciences \\
$^{2}$School of Artificial Intelligence, University of Chinese Academy of Sciences}
\begin{document}
\maketitle

\begin{abstract}
Despite recent advances in Aerial Vision-and-Language Navigation (VLN), single Unmanned Aerial Vehicle (UAV) search-and-rescue in outdoor scenarios remains challenging due to partial observability, making cross-altitude collaboration a promising direction.
However, existing collaborative navigation methods rely on precise target localization and rigid goal-dispatching paradigms, ignoring high-altitude observation uncertainty and suppressing low-altitude exploration autonomy.
To address these issues, we introduce \textbf{\modelname{}}, a training-free cross-altitude collaboration framework that enables robust navigation via \textbf{B}elief-\textbf{G}uided \textbf{C}oupling.
Specifically, the high-altitude UAV maintains a Global Belief Memory encoding task-conditioned priors, which the low-altitude agent incorporates into decision-making via Local Exhaustion Memory (LEM), Local Attraction Augmentation (LAA), and Directional Belief Prior (DBP) as an advisory bias rather than a rigid command.
Extensive experiments on ESARBench show that \modelname{} consistently outperforms state-of-the-art single-agent and collaborative methods.
Code is available at \href{https://github.com/uestchjw/BGC-Nav}{\textcolor{purple}{https://github.com/uestchjw/BGC-Nav}}.

\end{abstract}


\section{Introduction}
\label{sec:intro}

Aerial Vision-and-Language Navigation (VLN) enables Unmanned Aerial Vehicles (UAVs) to autonomously identify and locate targets based on language instructions, with applications such as disaster response, environmental monitoring, and security patrol~\cite{zhou2026memory, qi2026parse, AerialVLA, AerialVLN, Navagent, CoDrone, zhang2026your, tian2025uavs}.
Single-UAV search-and-rescue, however, remains challenging in large-scale outdoor scenarios due to cluttered surroundings and partial observability~\cite{ESARBench, DisasterBench, OnFly, xue2026deliberate, FSD-VLN, lin2025openvln, jiang2025longfly, jiang2026airnavigation}.
Introducing high-altitude UAVs to complement low-altitude agents provides a broader field of view and enhanced situational awareness~\cite{zeng2026can, zhang2026air}, making cross-altitude multi-UAV collaboration a promising paradigm to improve navigation performance.

\par
However, existing collaborative navigation methods~\cite{AeroDuo, CoNav-UAV, yu2026towards} exhibit two key limitations: (1) relying on precise target localization from high-altitude UAVs, which can be unreliable in complex environments due to perception degradation and spatial ambiguity, and (2) depending on a naive "dispatch-and-reach" cooperative paradigm, where high-altitude UAVs assign fixed coordinates for low-altitude agents to reach, thereby restricting the exploration autonomy of low-altitude UAVs.
We find that under inaccurate high-altitude guidance, existing collaborative methods fail to provide performance gains and can even underperform single-agent baselines.

\par
To address these issues, we propose \modelname{}, a training-free cross-altitude collaborative framework that leverages belief-guided coupling for robust navigation.
Our key insight lies in reformulating high-altitude guidance as an advisory bias rather than a rigid constraint, thereby leveraging global semantic priors while preserving the exploration autonomy of low-altitude agents.
Specifically, \modelname{} employs a two-tier framework where the high-altitude UAV incrementally constructs a Global Belief Memory (GBM), serving as a task-conditioned field that provides global semantic priors.
The low-altitude agent queries and exhaustion-discounts the GBM before incorporating it into action scoring via Local Attraction Augmentation and Directional Belief Prior, balancing global priors and egocentric exploration.

\par
To validate \modelname{}, we conduct extensive experiments on ESARBench, a challenging UAV-based search-and-rescue benchmark. Experimental results show that our method consistently outperforms existing navigation baselines and enhances performance across diverse backbones as a plug-and-play module.
Our main contributions are as follows:
\begin{itemize}[leftmargin=*]
    \item We identify critical limitations in existing cross-altitude collaborative navigation methods, where rigid command dispatch ignores high-altitude observation uncertainty and restricts low-altitude navigation autonomy.
    \item We introduce \modelname{}, a cross-altitude collaborative framework that achieves robust navigation via belief-guided coupling, which integrates high-altitude global belief memory into low-altitude decision-making through local attraction augmentation and directional belief prior.
    \item Extensive experiments demonstrate that \modelname{} consistently outperforms existing single-agent and collaborative navigation baselines, exhibiting strong robustness and plug-and-play adaptability.
\end{itemize}

\begin{figure*}[htbp]
  \centering
  \includegraphics[width=\linewidth, keepaspectratio]{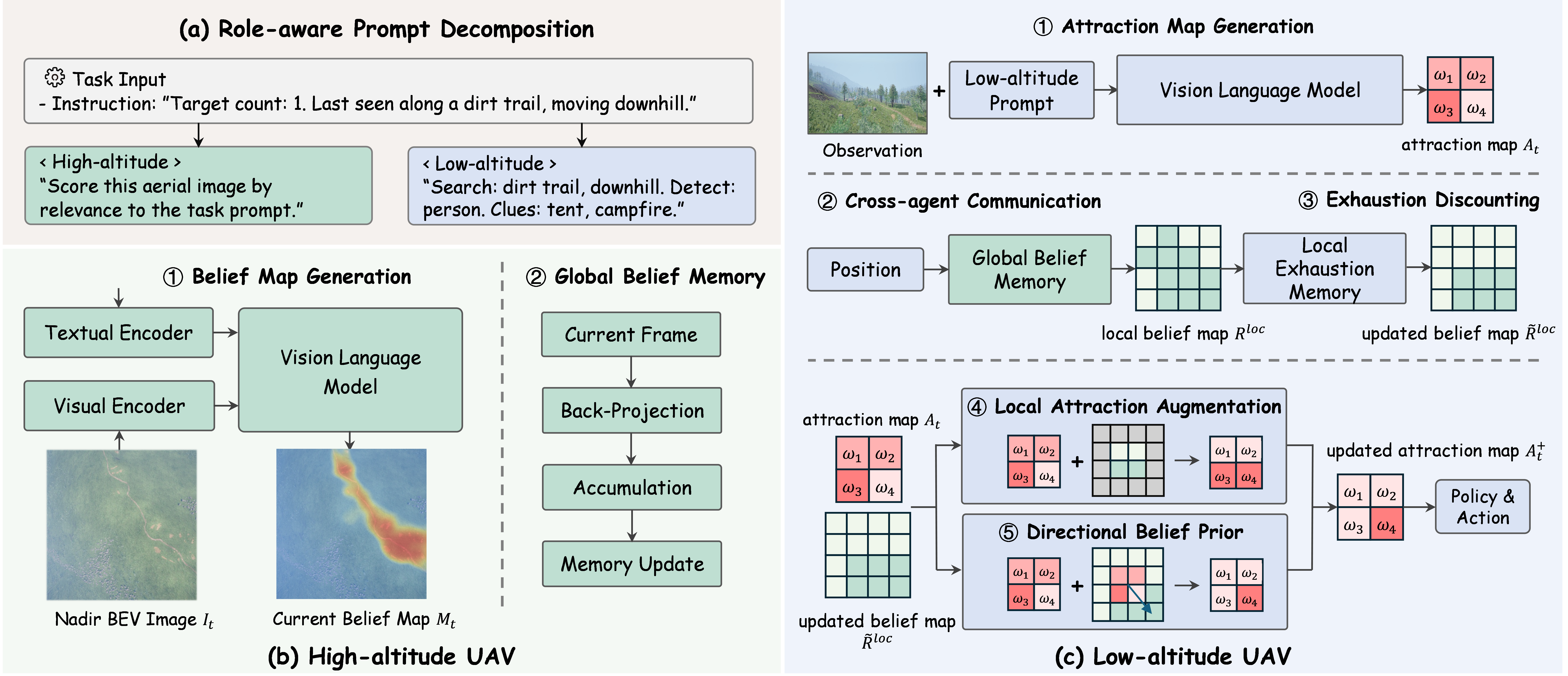}
\caption{Overview of \modelname{}. Following role-specific instruction decomposition, the high-altitude UAV maintains a Global Belief Memory encoding task-conditioned priors. The low-altitude UAV queries the Global Belief Memory to extract a local belief map, discounts it via a Local Exhaustion Memory, and modulates its attraction map using Local Attraction Augmentation and Directional Belief Prior for subsequent decision-making.}
  \label{figure:overview}
\end{figure*}

\section{Method}
\label{sec:method}

\subsection{Problem Formulation and Overview}
We consider UAV-based embodied search-and-rescue in an outdoor environment $\mathcal{E}$ with no prior terrain map. A mission specifies $K \ge 1$ victims $\{g_k\}_{k=1}^{K}$ and a natural-language prompt $\ell$ conveying the general area in which the victims were last seen. Physical clues (e.g., tents, backpacks) are scattered throughout $\mathcal{E}$.
The deployed UAVs must explore $\mathcal{E}$ and report the position of each $g_k$ within a fixed time budget $T$.

The overall architecture of \modelname{} is illustrated in Fig.~\ref{figure:overview}. We first decompose the task prompt $\ell$ into role-specific instructions. The high-altitude UAV constructs a global belief memory encoding task-conditioned relevance, which the low-altitude UAV dynamically incorporates via belief-guided coupling to inform decision-making.

\subsection{High-Altitude UAV}
Despite broad spatial coverage, high-altitude UAVs often struggle with accurate target localization in complex scenarios. We thus construct a task-conditioned field that identifies regions of interest and provides advisory semantic guidance to the low-altitude agent, rather than rigid waypoints.
\par
At each time step $t$, the high-altitude UAV captures a bird's-eye-view (BEV) image $I_t$, and a vision-language model $f_{\mathrm{vlm}}$ scores it conditioned on the high-altitude prompt $\ell_{\mathrm{H}}$, producing a frame-level belief map
\begin{equation}
M_t = f_{\mathrm{vlm}}(I_t, \ell_{\mathrm{H}}) \in [0,1]^{G \times G},
\label{eq:localbelief}
\end{equation}
where $G$ denotes the spatial resolution of the belief map. Throughout the exploration horizon, the high-altitude UAV persistently advances into unexplored regions. The Global Belief Memory (GBM) is incrementally updated by projecting each frame-level belief map $M_t$ onto a world-aligned grid (with cell $i$ located at $c_i$) and averaging it with historical observations. This yields a per-cell belief $R_i \in [0,1]$, with a low prior $r_0$ assigned to unexplored cells.

\subsection{Low-Altitude UAV}
We instantiate the low-altitude navigation framework with APEX \cite{APEX}, though the proposed coupling mechanism can seamlessly generalize to alternative methods.

At time step $t$, the low-altitude UAV is located at $p_t$ and, based on its own observation and the task prompt $\ell_{\mathrm{L}}$, generates an egocentric 3D attraction map $A_t$, whose values represent the attraction score of candidate regions.
We incorporate the global semantic prior from the high-altitude UAV by adding a bias to this attraction map.

\textbf{Local belief map.} At each step, the low-altitude UAV queries the GBM to retrieve a local belief map $R^{\mathrm{loc}}$ centered at its position $p_t$ with radius $W$. Since $W$ exceeds the spatial extent of $A_t$, $R^{\mathrm{loc}}$ provides semantic priors for regions beyond the local field of view.

\textbf{Local Exhaustion Memory (LEM).} To prevent repetitive exploration of previously searched regions, the UAV modulates $R^{\mathrm{loc}}$ using a Local Exhaustion Memory (LEM) derived from its trajectory history. Specifically, for each cell $c_i$ in the local belief map, the visitation count $n_i$ accumulates whenever the UAV operates within its neighborhood $\rho$:
\begin{equation}
n_i \leftarrow n_i + \mathbbm{1}[\lVert c_i - p_t\rVert \le \rho].
\end{equation}
The belief score at cell $i$ is then discounted via an exponential decay function:
\begin{equation}
\tilde{R}^{\mathrm{loc}}_i = r_0 + \big(R^{\mathrm{loc}}_i - r_0\big)\, e^{-n_i/\tau_o},
\label{eq:memmask}
\end{equation}
where $r_0$ denotes the belief floor and $\tau_o$ controls the attenuation rate.

The updated local belief map $\tilde{R}^{\mathrm{loc}}$ modifies the attraction map $A_t$ through two complementary modules: Local Attraction Augmentation and Directional Belief Prior. This bias is applied only during candidate action evaluation and is not saved to $A_t$, preventing state accumulation over time. Furthermore, if $\max A_t$ exceeds a detection threshold $\tau_{\mathrm{det}}$, indicating that the UAV has directly detected the target or a visual cue, the coupling is skipped to retain the original $A_t$.

\textbf{Local Attraction Augmentation (LAA).} For each horizontal cell $(j,k)$ in $A_t$ at world coordinates $(x_{jk},y_{jk})$, the LAA module adds an attraction offset proportional to the discounted local belief:
\begin{equation}
b^{\mathrm{loc}}_{jk} = \mathrm{clip}\big(\gamma_\ell\,(\tilde{R}^{\mathrm{loc}}(x_{jk},y_{jk}) - r_0),\,0,\,\beta_\ell\big),
\label{eq:local}
\end{equation}
where $\gamma_\ell$ is a scaling gain and $\beta_\ell$ limits the maximum boost to amplify high-belief candidates within $A_t$.

\textbf{Directional Belief Prior (DBP).} The DBP module provides guidance toward high-belief regions beyond $A_t$. It groups cells satisfying $\tilde{R}^{\mathrm{loc}}_i \ge \tau_m$ into connected components and selects $\mathcal{C}_m^*$ maximizing the distance-attenuated belief mass:
\begin{equation}
\mathcal{C}_m^* = \arg\max_{\mathcal{C}_m} \frac{\sum_{i\in\mathcal{C}_m}\tilde{R}^{\mathrm{loc}}_i}{1+\lVert \bar c_m - p_t\rVert / D},
\end{equation}
where $\bar c_m$ is the belief-weighted centroid and $D$ is a distance scale that attenuates distant components. The bearing from $p_t$ to $\bar c_m$ is updated each step and serves as a directional prior rather than an explicit waypoint. Grid cells within a wedge of half-angle $\phi$ aligned with this bearing receive:
\begin{equation}
b^{\mathrm{dir}}_{jk} = (1-v_t)\,\min\big(\beta_w,\ \gamma_w\,\alpha\, w_{jk}\big),
\label{eq:lean}
\end{equation}
where $w_{jk}\in[0,1]$ tapers with angular offset and grows with distance. The binary variable $v_t\in\{0,1\}$ suppresses the bias when line-of-sight obstacles block the path, while $\alpha\in[\alpha_{\min},1]$ fades the influence as the UAV approaches $\bar c_m$. Parameters $\gamma_w$ and $\beta_w$ denote the gain and upper saturation bound.

The two bias terms are summed, saturated at $\beta$, and enforced as an element-wise lower bound on $A_t$:
\begin{equation}
A^{+}_t = \max\big(A_t,\ \mathrm{clip}(b^{\mathrm{loc}}+b^{\mathrm{dir}},\,0,\,\beta)\big),
\label{eq:combine}
\end{equation}
enabling the low-altitude UAV to select actions via its unmodified policy. Because $\beta$ never overrides confirmed local detections, the bias serves solely to reorder competing candidate actions.

\begin{table}[t]
  \centering
  \setlength{\abovecaptionskip}{0.1cm}
  \setlength{\belowcaptionskip}{0.1cm}
  \caption{Performance comparison on ESARBench across ESAR-short ($<$200\,m) and ESAR-long ($\geq$200\,m) subsets, split by start-to-target distance.}
  \small
  \setlength{\tabcolsep}{3pt}
  \resizebox{\linewidth}{!}{
  \begin{tabular}{@{}cc*{6}{c}@{}}
  \toprule
  \multirow{2}{*}{\textbf{Mode}} & \multirow{2}{*}{\textbf{Method}}
  & \multicolumn{3}{c}{\textbf{ESAR-short}} & \multicolumn{3}{c}{\textbf{ESAR-long}}\\
  \cmidrule(lr){3-5} \cmidrule(lr){6-8}
  & & \textbf{SR$\uparrow$} & \textbf{OSR@40$\uparrow$} & \textbf{RS$\uparrow$}
  & \textbf{SR$\uparrow$} & \textbf{OSR@40$\uparrow$} & \textbf{RS$\uparrow$} \\
  \midrule
  \multirow{5}{*}{Single-UAV}
  & FBE \cite{FBE}   & 9.86 & 9.72 & 11.30 & 4.34 & 5.03 & 10.03 \\
  & VLFM \cite{VLFM} & 12.82 & 14.74 & 12.87 & 5.70 & 6.56 & 9.97 \\
  & SPF \cite{SPF} & 12.19 & 11.37 & 9.09 & 3.31 & 3.01 & 8.24 \\
  & UniGoal \cite{yin2025unigoal} & 4.65 & 6.46 & 8.47 & 3.13 & 1.89 & 7.14 \\
  & APEX \cite{APEX} & 12.61 & 18.20 & 14.68 & 4.52 & 2.58 & 10.93 \\
\midrule
\multirow{4}{*}{Dual-UAV}
& CoNav-UAV \cite{CoNav-UAV} & 10.57 & 7.45 & 14.59 & 6.10 & 4.06 & 12.38 \\
& AeroDuo \cite{AeroDuo} & 12.84 & 11.99 & 15.39 &  10.95 & \textbf{11.82} & 13.97 \\
& Sap-UAV \cite{Sap-UAV} & 12.23 & 10.60 & 15.59 &  9.12 & 6.30 & 14.13 \\
& \cellcolor{lightgray3}\textbf{\modelname{}}
& \cellcolor{lightgray3}\textbf{14.56} & \cellcolor{lightgray3}\textbf{19.81} & \cellcolor{lightgray3}\textbf{17.17}
& \cellcolor{lightgray3}\textbf{13.24} & \cellcolor{lightgray3}9.85 & \cellcolor{lightgray3}\textbf{15.49} \\
\bottomrule
  \end{tabular}}

  \label{table:overall}
\end{table}

\section{Experiments}
\label{sec:experiments}

\subsection{Experimental Settings}
\label{sec:setup}

\textbf{Datasets.} We evaluate \modelname{} on ESARBench \cite{ESARBench}, a UAV embodied search and rescue benchmark featuring four large-scale, high-fidelity environments across 600 tasks. Based on start-to-target distance, we partition the dataset into ESAR-short ($< 200\,\text{m}$) and ESAR-long ($\ge 200\,\text{m}$).
\par
\textbf{Evaluation metrics.}
Following \cite{ESARBench}, we report Success Rate (SR) for reaching within 15\,\text{m} of the target, Clue Discovery Score (CDS) for clue localization, and Rescue Score (RS) to balance task completion, success, and time efficiency.
Additionally, we adopt Oracle Success Rate (OSR@$k$), measuring the percentage of episodes whose trajectories come within $k$\,\text{m} of a ground-truth target.

\definecolor{darkgreen}{RGB}{0,140,60}
\definecolor{darkred}{RGB}{200,30,30}
\begin{table}[t]
\centering
\caption{Performance comparison under varying levels of high-altitude guidance reliability.}
\vspace{-6pt}
\label{table:guidance_quality}
\resizebox{\columnwidth}{!}{%
\begin{tabular}{clllll}
\toprule
\textbf{Reliability} & \textbf{Method} & \textbf{SR$\uparrow$} & \textbf{RS$\uparrow$} & \textbf{OSR@30$\uparrow$} & \textbf{OSR@60$\uparrow$} \\
\midrule
\multirow{3}{*}{Low} & APEX~\cite{APEX} & \textbf{3.70} & 10.60 & 3.70 & 7.41 \\
 & AeroDuo~\cite{AeroDuo} & 1.85{\scriptsize\color{darkred}$_{1.85\downarrow}$} & 8.71{\scriptsize\color{darkred}$_{1.89\downarrow}$} & 0.10{\scriptsize\color{darkred}$_{3.60\downarrow}$} & 0.93{\scriptsize\color{darkred}$_{6.48\downarrow}$} \\
 & \cellcolor{lightgray3}\modelname{} & \cellcolor{lightgray3}3.09{\scriptsize\color{darkred}$_{0.61\downarrow}$} & \cellcolor{lightgray3}\textbf{10.89}{\scriptsize\color{darkgreen}$_{0.29\uparrow}$} & \cellcolor{lightgray3}\textbf{4.32}{\scriptsize\color{darkgreen}$_{0.62\uparrow}$} & \cellcolor{lightgray3}\textbf{10.19}{\scriptsize\color{darkgreen}$_{2.78\uparrow}$} \\
\midrule
\multirow{3}{*}{Mid} & APEX~\cite{APEX} & \textbf{13.87} & \textbf{16.60} & 3.65 & 14.37 \\
 & AeroDuo~\cite{AeroDuo} & 11.08{\scriptsize\color{darkred}$_{2.79\downarrow}$} & 15.17{\scriptsize\color{darkred}$_{1.43\downarrow}$} & 3.15{\scriptsize\color{darkred}$_{0.50\downarrow}$} & 14.59{\scriptsize\color{darkgreen}$_{0.22\uparrow}$} \\
 & \cellcolor{lightgray3}\modelname{} & \cellcolor{lightgray3}12.03{\scriptsize\color{darkred}$_{1.84\downarrow}$} & \cellcolor{lightgray3}15.76{\scriptsize\color{darkred}$_{0.84\downarrow}$} & \cellcolor{lightgray3}\textbf{7.66}{\scriptsize\color{darkgreen}$_{4.01\uparrow}$} & \cellcolor{lightgray3}\textbf{16.53}{\scriptsize\color{darkgreen}$_{2.16\uparrow}$} \\
\midrule
\multirow{3}{*}{High} & APEX~\cite{APEX} & 8.90 & 14.91 & 10.48 & 22.99 \\
 & AeroDuo~\cite{AeroDuo} & 23.78{\scriptsize\color{darkgreen}$_{14.88\uparrow}$} & 20.12{\scriptsize\color{darkgreen}$_{5.21\uparrow}$} & \textbf{17.29}{\scriptsize\color{darkgreen}$_{6.81\uparrow}$} & 34.20{\scriptsize\color{darkgreen}$_{11.21\uparrow}$} \\
 & \cellcolor{lightgray3}\modelname{} & \cellcolor{lightgray3}\textbf{25.44}{\scriptsize\color{darkgreen}$_{16.54\uparrow}$} & \cellcolor{lightgray3}\textbf{21.77}{\scriptsize\color{darkgreen}$_{6.86\uparrow}$} & \cellcolor{lightgray3}16.63{\scriptsize\color{darkgreen}$_{6.15\uparrow}$} & \cellcolor{lightgray3}\textbf{37.43}{\scriptsize\color{darkgreen}$_{14.44\uparrow}$} \\
\bottomrule
\end{tabular}%
}
\end{table}

\textbf{Implementation details.} We set the high-altitude belief grid resolution to $25\,\text{m}$ per cell. The attraction map covers a $200\,\text{m} \times 200\,\text{m} \times 50\,\text{m}$ volume with a $250\,\text{m}$ radius local belief window ($\rho{=}60\,\text{m}$, $\tau_o{=}12$ steps for LEM). The coupling bias is capped at $\beta = 0.65$, between the stand-down gate ($\tau_{\mathrm{det}}=0.3$) and the confirmed detection threshold ($0.90$).
All experiments are conducted on NVIDIA RTX A6000 GPUs.

\begin{figure}[t]
    \centering
    \begin{subfigure}[b]{0.49\linewidth}
        \centering
        \includegraphics[width=\linewidth]{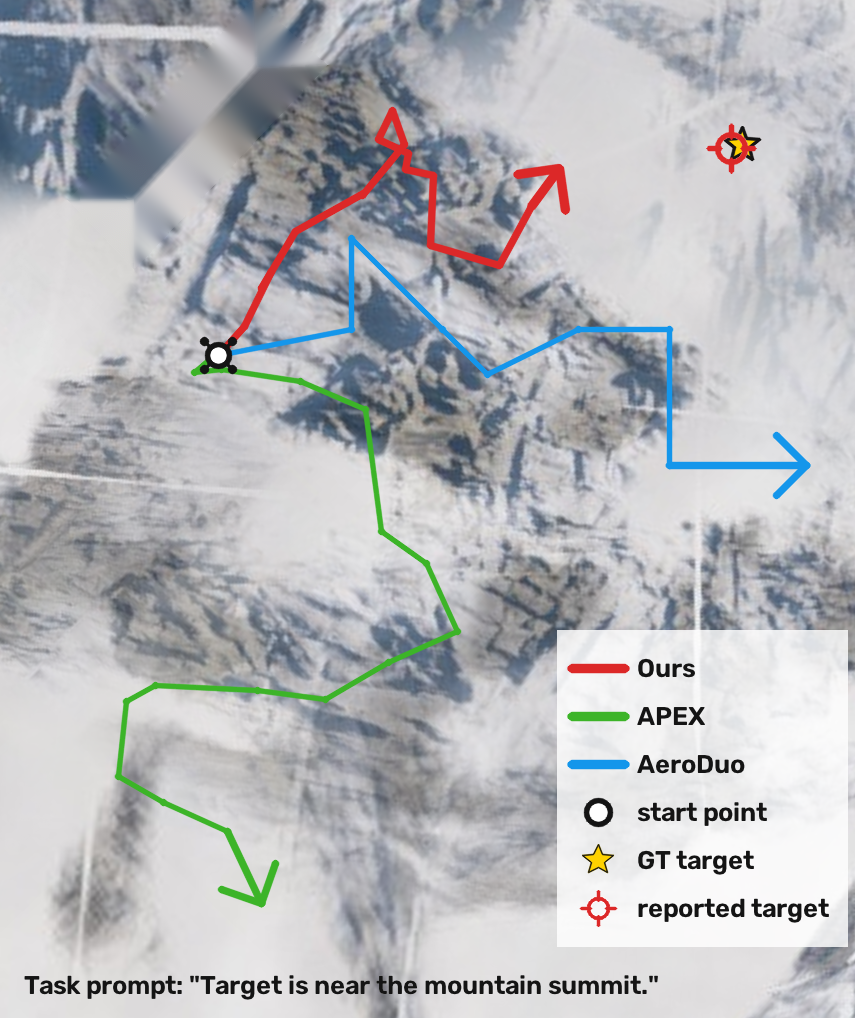}
        \caption{Orthographic map}
        \label{fig:task434-bev}
    \end{subfigure}
    \hfill
    \begin{subfigure}[b]{0.49\linewidth}
        \centering
        \includegraphics[width=\linewidth]{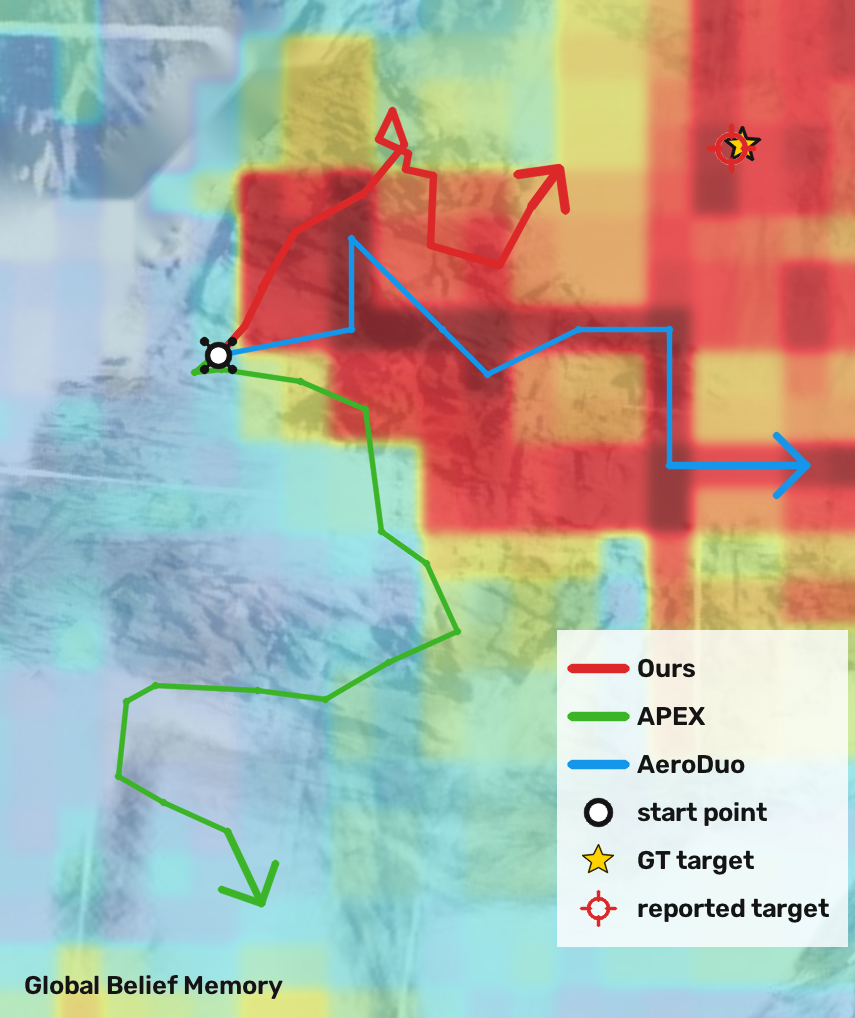}
        \caption{Global belief memory}
        \label{fig:task434-belief}
    \end{subfigure}
    \caption{Qualitative comparison of navigation trajectories. Unlike the dispatch-and-reach collaborative method (AeroDuo) that rigidly enforces target waypoints, \modelname{} maintains robust target search performance by balancing global semantic guidance and exploration autonomy.}

    \label{fig:Qualitative Evaluation}
\end{figure}

\begin{table}[t]
\centering
\caption{Performance comparison across different belief-map generation models.}
\vspace{-6pt}
\label{tab:belief_source}
\resizebox{\columnwidth}{!}{%
\begin{tabular}{lllll}
\toprule
\textbf{Belief Source} & \textbf{Method} & \textbf{SR$\uparrow$} & \textbf{RS$\uparrow$} & \textbf{OSR@40$\uparrow$} \\
\midrule
\multirow{2}{*}{CLIPSeg~\cite{CLIPSeg}} & AeroDuo~\cite{AeroDuo} & 6.67 & 12.70 & 8.30 \\
 & \cellcolor{lightgray3}\modelname{} & \cellcolor{lightgray3}12.82{\scriptsize\color{darkgreen}$_{6.15\uparrow}$} & \cellcolor{lightgray3}15.06{\scriptsize\color{darkgreen}$_{2.36\uparrow}$} & \cellcolor{lightgray3}8.97{\scriptsize\color{darkgreen}$_{0.67\uparrow}$} \\
\midrule
\multirow{2}{*}{Claude Sonnet 5~\cite{Claude}} & AeroDuo~\cite{AeroDuo} & 10.95 & 13.97 & 11.82 \\
 & \cellcolor{lightgray3}\modelname{} & \cellcolor{lightgray3}13.24{\scriptsize\color{darkgreen}$_{2.29\uparrow}$} & \cellcolor{lightgray3}15.49{\scriptsize\color{darkgreen}$_{1.52\uparrow}$} & \cellcolor{lightgray3}9.85{\scriptsize\color{darkred}$_{1.97\downarrow}$} \\
\bottomrule
\end{tabular}%
}
\end{table}

\subsection{Quantitative Evaluation}

\textbf{Comparison with state-of-the-art.} Table~\ref{table:overall} compares \modelname{} against existing navigation approaches. Our \modelname{} consistently outperforms both single-agent and collaborative baselines, particularly on the challenging ESAR-long subset.
Notably, \modelname{} surpasses Sap-UAV~\cite{Sap-UAV}, which coordinates two low-altitude UAVs via spatial partitioning, demonstrating that high-altitude semantic priors effectively guide low-altitude search.

\par
\textbf{Robustness to guidance reliability.}
Table~\ref{table:guidance_quality} evaluates performance across different levels of high-altitude guidance reliability, measured by the rank of the ground-truth cell in the belief map. Under reliable guidance, both the dispatch-and-reach method AeroDuo~\cite{AeroDuo} and ours \modelname{} outperform the single-agent baseline APEX~\cite{APEX}. As reliability degrades, however, AeroDuo suffers a drastic performance drop, whereas \modelname{} remains competitive by balancing global priors and ego-centric autonomy.

\textbf{Robustness to belief sources.} Table~\ref{tab:belief_source} compares performance across different belief-map generation backbones. Across all models, \modelname{} consistently outperforms AeroDuo~\cite{AeroDuo}. Notably, as belief accuracy decreases (transitioning from Claude to CLIPSeg), AeroDuo degrades significantly, whereas \modelname{} retains high performance, demonstrating the effectiveness of our belief-guided coupling.

\par
\textbf{Plug-and-play generalization.} We evaluate our belief-guided coupling as a plug-and-play module across diverse navigation methods. As presented in Table~\ref{table:generalization}, the proposed mechanism yields consistent performance gains across all backbones, demonstrating strong cross-model generalization.

\subsection{Qualitative Analysis}
Fig.~\ref{fig:Qualitative Evaluation} visualizes navigation trajectories across different methods. When high-altitude guidance is imprecise, dispatch-and-reach baselines strictly commit to false coordinates. In contrast, \modelname{} achieves robust target search by effectively balancing global semantic priors with low-altitude exploratory autonomy.

\begin{table}[t]
\centering
\caption{Integration of our training-free belief-guided coupling mechanism into existing exploration methods.}
\vspace{-6pt}
\label{table:generalization}
\resizebox{\columnwidth}{!}{%
\begin{tabular}{llllll}
\toprule
\textbf{Method} & \textbf{SR}$\uparrow$ & \textbf{OSR@30}$\uparrow$ & \textbf{OSR@60}$\uparrow$ & \textbf{CDS}$\uparrow$ & \textbf{RS}$\uparrow$ \\
\midrule
FBE~\cite{FBE} & 4.34 & 2.01 & 8.94 & 3.10 & 10.03 \\
\rowcolor{lightgray3}
\quad + Ours & 12.82{\scriptsize\color{darkgreen}$_{8.48\uparrow}$} & 4.03{\scriptsize\color{darkgreen}$_{2.02\uparrow}$} & 17.31{\scriptsize\color{darkgreen}$_{8.37\uparrow}$} & 6.42{\scriptsize\color{darkgreen}$_{3.32\uparrow}$} & 13.21{\scriptsize\color{darkgreen}$_{3.18\uparrow}$} \\
\midrule
VLFM~\cite{VLFM} & 5.70 & 4.19 & 8.76 & 4.10 & 9.97 \\
\rowcolor{lightgray3}
\quad + Ours & 6.93{\scriptsize\color{darkgreen}$_{1.23\uparrow}$} & 8.42{\scriptsize\color{darkgreen}$_{4.23\uparrow}$} & 9.49{\scriptsize\color{darkgreen}$_{0.73\uparrow}$} & 4.41{\scriptsize\color{darkgreen}$_{0.31\uparrow}$} & 13.42{\scriptsize\color{darkgreen}$_{3.45\uparrow}$} \\
\midrule
APEX~\cite{APEX} & 4.52 & 2.20 & 7.35 & 2.36 & 10.93 \\
\rowcolor{lightgray3}
\quad + Ours & 13.24{\scriptsize\color{darkgreen}$_{8.72\uparrow}$} & 6.21{\scriptsize\color{darkgreen}$_{4.01\uparrow}$} & 14.05{\scriptsize\color{darkgreen}$_{6.70\uparrow}$} & 4.05{\scriptsize\color{darkgreen}$_{1.69\uparrow}$} & 15.49{\scriptsize\color{darkgreen}$_{4.56\uparrow}$} \\
\bottomrule
\end{tabular}%
}
\end{table}

\begin{table}[t]
\centering
\caption{Ablation study of LAA (Local Attraction Augmentation), DBP (Directional Belief Prior), and LEM (Local Exhaustion Memory) components.}
\vspace{-6pt}
\label{table:ablation}
\resizebox{\columnwidth}{!}{%
\begin{tabular}{lccccc}
\toprule
\textbf{Method} & \textbf{SR$\uparrow$} & \textbf{OSR@30$\uparrow$} & \textbf{OSR@60$\uparrow$} & \textbf{CDS$\uparrow$} & \textbf{RS$\uparrow$} \\
\midrule
Ours w/o LAA & 7.31 & 6.08 & 11.29 & 3.98 & 13.56 \\
Ours w/o DBP & 7.69 & 6.21 & 7.95 & 2.00 & 12.77 \\
Ours w/o LEM & 8.39 & 4.17 & 10.42 & 4.02 & 13.42 \\
\rowcolor{lightgray3}
Ours & \textbf{13.24} & \textbf{6.21} & \textbf{14.05} & \textbf{4.05} & \textbf{15.49} \\
\bottomrule
\end{tabular}%
}
\end{table}

\subsection{Ablation Study}

To validate the effectiveness of our proposed components, we conduct ablation studies as shown in Table~\ref{table:ablation}.
Removing Local Attraction Augmentation (LAA) causes the largest SR drop, highlighting its role in local target search. Without Directional Belief Prior (DBP), OSR@60 decreases markedly, indicating that directional priors extend search coverage toward distant targets. Moreover, Local Exhaustion Memory (LEM) brings consistent gains by discounting previously searched regions and suppressing redundant exploration.

\section{Conclusion}

This paper fundamentally rethinks cooperative strategies in existing multi-UAV navigation methods. We propose \modelname{}, a training-free framework for cross-altitude collaborative navigation via belief-guided coupling, which leverages global semantic priors while preserving exploration autonomy. The high-altitude UAV maintains a Global Belief Memory of task-conditioned priors, which the low-altitude agent integrates into action scoring via advisory biasing. Experimental results demonstrate the effectiveness of our approach.

\clearpage
\bibliographystyle{IEEEbib}
\bibliography{refs}

\end{document}